\documentclass[runningheads]{llncs}

\usepackage{accv}

\usepackage{accvabbrv}

\usepackage{graphicx}
\usepackage{booktabs}
\usepackage{multirow}
\usepackage{array}
\usepackage{rotating}
\usepackage{tabularx}
\usepackage{url}
\newcolumntype{P}[1]{>{\centering\arraybackslash}p{#1}}
\newcolumntype{Y}{>{\centering\arraybackslash}X}
\usepackage{amssymb}
\DeclareMathOperator*{\argmax}{arg\,max}
\DeclareMathOperator*{\argmin}{arg\,min}

\usepackage[accsupp]{axessibility}  % Improves PDF readability for those with disabilities.

\usepackage{hyperref}

\usepackage{orcidlink}

\begin{document}

\title{RayEmb: Arbitrary Landmark Detection in X-Ray Images Using Ray Embedding Subspace} 

\titlerunning{RayEmb}

\author{Pragyan Shrestha\inst{1}\orcidlink{0009-0000-8276-4116} \and
Chun Xie\inst{1}\orcidlink{0000-0003-4936-7404} \and
Yuichi Yoshii\inst{2}\orcidlink{0000-0003-1447-4664} \and
Itaru Kitahara\inst{1}\orcidlink{0000-0002-5186-789X}}

\authorrunning{P.~Shrestha et al.}

\institute{University of Tsukuba, Tsukuba, Ibaraki, Japan 
\email{shrestha.pragyan@image.iit.tsukuba.ac.jp}
\email{\{xiechun,kitahara\}@ccs.tsukuba.ac.jp}
\and
Tokyo Medical University, Ami, Ibaraki, Japan\\
\email{yyoshii@tokyo-med.ac.jp}
}

\maketitle

\begin{abstract}
Intra-operative 2D-3D registration of X-ray images with pre-operatively acquired CT scans is a crucial procedure in orthopedic surgeries. Anatomical landmarks pre-annotated in the CT volume can be detected in X-ray images to establish 2D-3D correspondences, which are then utilized for registration. However, registration often fails in certain view angles due to poor landmark visibility. We propose a novel method to address this issue by detecting arbitrary landmark points in X-ray images. Our approach represents 3D points as distinct subspaces, formed by feature vectors (referred to as ray embeddings) corresponding to intersecting rays. Establishing 2D-3D correspondences then becomes a task of finding ray embeddings that are close to a given subspace, essentially performing an intersection test. Unlike conventional methods for landmark estimation, our approach eliminates the need for manually annotating fixed landmarks. We trained our model using the synthetic images generated from CTPelvic1K CLINIC dataset, which contains 103 CT volumes, and evaluated it on the DeepFluoro dataset, comprising real X-ray images. Experimental results demonstrate the superiority of our method over conventional methods. The code is available at \url{https://github.com/Pragyanstha/rayemb}.
\end{abstract}

\section{Introduction}
\label{sec:intro}
Intra-operative 2D-3D registration of X-ray images with pre-operatively obtained CT scans is a widely used technique in orthopedic surgeries. In a typical surgical setting, a fluoroscopy device captures an X-ray image of the target structure within the patient, who is positioned on the operating table. The X-ray image is a projection of the target structure based on the geometric setup of the device, making it challenging for surgeons to visualize the actual 3D anatomy. 2D-3D registration enables overlaying the 3D model, derived from the pre-operative CT scan, onto the X-ray image. Additionally, surgical planning data, such as the placement of implants and pedicle screws, can be visualized similarly through registration. In clinical practice, procedures like total hip arthroplasty, total knee arthroplasty, osteosynthesis, and other trauma surgeries utilize 2D-3D registration to understand the spatial arrangement of implants and guide surgical instruments during the operation.\cite{Bradley2019-qv,Wada2017-na,Yoshii2017-rr}. 
\begin{figure}[t]
  \centering
  \includegraphics[width=1.0\textwidth]{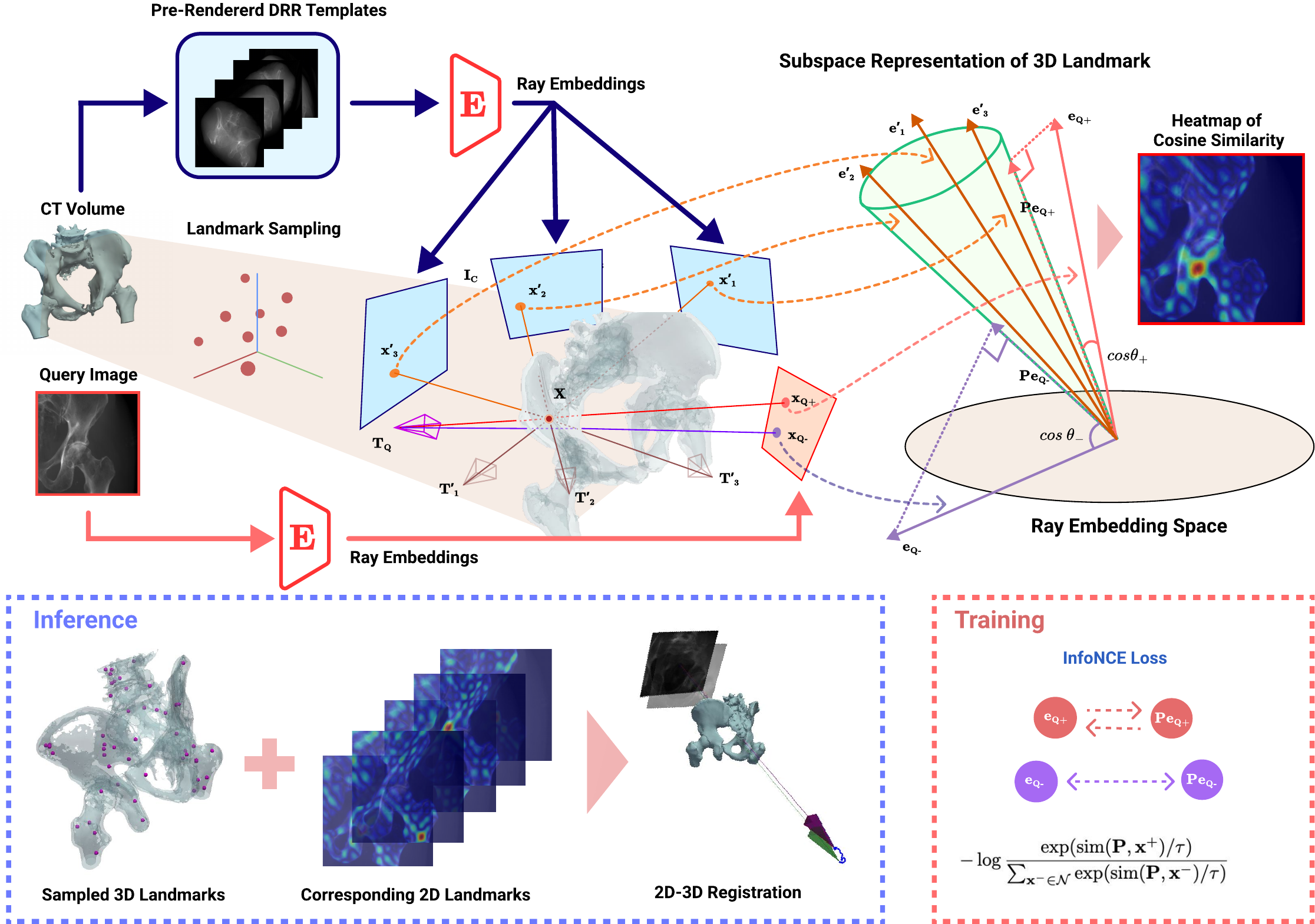}
  \caption{
Given a query X-ray image and its corresponding CT volume, 3D landmark points within the volume are randomly sampled. For each sampled point, the feature vectors of its projected 2D points are collected to form a subspace spanned by these vectors. The feature vectors from the query X-ray image are then compared with their projections onto each subspace. This process generates a heatmap that exhibits strong activation near the 2D location corresponding to the projected 3D landmark.}
  \label{fig:overview}
\end{figure}
In its technical aspect, the problem of 2D-3D registration closely resembles 6-DoF object pose estimation. However, a notable difference lies in the field of view, specifically whether the target object is truncated. Many 6-DoF object pose estimation methods in the computer vision literature employ object detectors such as YOLO\cite{Redmon2016-ja}, SSD\cite{Liu2016-bw}, FasterRCNN\cite{Ren2015-jt} and RetinaNet\cite{Lin2017-wq}, prior to their pose estimation pipeline. This detection step is generally omitted in 2D-3D registration, especially when determining the pose of a target structure under partial visibility. Another major difference is the domain of the input; while it is usually RGB(D) in the literature, methods developed for RGB(D) images are not directly applicable to X-ray images due to fundamental differences in their image formation processes. Specifically, methods designed for RGB(D) images build upon the fact that the pixel intensities in a patch region provide information about the surface appearance of the corresponding object. Many studies have exploited this characteristics for establishing 2D-3D correspondences between the image points and the corresponding points on the surface\cite{Zakharov2019-qn,Hodan2020-ps,Haugaard2022-wk}. Similarly, template matching based approaches \cite{Labbe2020-ak,Nguyen2024-td} work based on similarity of patches across the template and query image. However, these underlying assumptions do not hold in transmissive imaging technique such as X-ray images, making 2D-3D registration a more challenging task. 

Recent advances in 2D-3D registration have shown significant progress by adopting learning based methods. One promising approach involves establishing 2D-3D correspondences by localizing predefined landmarks in the X-ray images\cite{Bier2018-ht,Grupp2020-xz,Esteban2019-zg}. These methods train a model to predict a heatmap of landmarks using keypoint-annotated images. During the registration process, the landmark annotations in the CT volume, combined with the predicted 2D heatmap, are used to establish the 2D-3D correspondences, which then serve as input for perspective-n-point (PnP) algorithms with random sample and consensus (RANSAC) to filter outliers. Although this strategy achieves high registration accuracy on successfully registered images, it often fails when landmark visibility is limited due to partial views of the structure. Additionally, annotating landmarks in the CT volume is not trivial, as it requires domain expertise to precisely identify clinically meaningful anatomical landmarks specific to the target structure. The reliance on these annotated landmarks during inference further complicates use in emergency situations, where a quick setup is essential.

To overcome the limitations of current landmark-based registration and enhance applicability across patients, we propose a novel method for localizing the 2D projections of arbitrary 3D points within a specified CT volume. By leveraging the transmissive nature of X-ray images, our approach hypothesize that a patch in an X-ray image contains information about the corresponding ray and its attenuation distribution along that ray. Thus, a 3D point within a CT volume can be uniquely represented by a collection of rays passing through it. In other words, rays intersecting this point can be distinguished by analyzing the information contained in the associated pixels. Specifically, we train an encoder to learn ray embeddings such that embeddings of intersecting rays span a distinctive subspace, which can be used to identify whether a query ray embedding intersects at the same point. This differentiation from other rays not intersecting the point is mathematically realized using subspace and the similarity of a vector with its projection onto the subspace. While some previous works with similar objective have focused on establishing 3D-3D or 2D-2D correspondences \cite{Yan2022-ik,Liu2021-kn}. To the best of our knowledge, this is the first attempt at establishing 2D-3D correspondences of arbitrary landmark points.

In summary, our main contributions are as follows.
\begin{itemize}
    \item A novel method for establishing arbitrary landmark correspondences within the field of view, enabling 2D-3D registration even in difficult views with partial visibility.
    \item A sampling strategy for finding good 2D-3D correspondences through test time augmentation of selected templates.
    \item A self-supervised training strategy utilizing synthetic data generation, where landmark annotation by domain expert is not required.  
\end{itemize}

\section{Related Works}
\subsubsection{6DoF Object Pose Estimation.}
Various approaches have been proposed for solving 6DoF object pose estimation. Direct methods such as PoseNet\cite{Kendall2015-mk}, PoseCNN\cite{Xiang2017-ae}, Deep-6DPose\cite{Do2018-ke}, and EfficientPose\cite{Bukschat2020-kd} regress the rotation and translation vectors of detected objects. In contrast, methods like DeepIM\cite{Li2018-dm}, CosyPose\cite{Labbe2020-ak}, and MegaPose\cite{Labb-e2022-mm} estimate the pose updates iteratively by comparing the current rendering with the observed image. Indirect methods solve a proxy problem by finding correspondences, which are then used in PnP-RANSAC framework to derive the final pose. OSOP\cite{Shugurov2022-ge} establishes 2D-3D correspondences indirectly by find a matching template, with known pose, and establishing 2D-2D correspondences with it. Similarly, Gigapose\cite{Nguyen2024-td} first finds the out-of-plane rotation by matching templates and estimates the remaining 4-DoF using patch correspondences. BB8\cite{Rad2017-kb} locates the 2D projections of the corners of 3D bounding boxes to establish correspondences. To address the challenge of truncated objects where keypoints may lie outside the image, PVNet\cite{Peng2019-ie} regresses vectors that point towards the keypoints, determining the final location through a voting mechanism. In dense prediction-based methods, Pix2Pose\cite{Park2019-oe} regresses the 3D coordinates of corresponding pixels to establish dense 2D-3D correspondences. Inspired by DensePose\cite{Guler2018-ee}, which maps all  pixels of a human RGB image to the 3D surface of the human body, DPOD\cite{Zakharov2019-qn} adopts a different approach by estimating UV texture maps of the object. Similarly, EPOS\cite{Hodan2020-ps} predicts multiple surface fragment coordinates for a pixel, allowing for the modeling of pose ambiguities due to partial symmetries in the object. Building on EPOS, Surfemb\cite{Haugaard2022-wk} estimates a continuous distribution over the object surface and uses this distribution for pose scoring and refinement. GDRNet\cite{Wang2021-eb} employs a hybrid approach by regressing dense geometric features for establishing 2D-3D correspondences and uses patch-PnP network for direct pose regression, rather than the conventional PnP-RANSAC framework.

\subsubsection{Learning based 2D-3D Registration.}
Traditionally, 2D-3D registration in medical imaging have incorporated two main approaches in clinical practice: image based optimization of digitally reconstructed radiographs (DRRs) from CT volumes against X-ray images \cite{Markelj2012-dc}, and the establishment of 2D-3D correspondences through pre-embedded fiducial landmarks visible in X-ray images \cite{George2011-ep}. Recent advancements in learning-based methods have aimed to overcome the limitations inherent in these traditional approaches. Some works have focused on extending the capture range of optimization schemes by iteratively updating pose estimates \cite{Gu2020-rx,Gopalakrishnan2024-yq}. These learning-based methods, which involve comparing DRRs with actual X-ray images to estimate pose differences, achieve better registration accuracy through further refinement via image similarity-based optimization. However, initial estimates often remain suboptimal compared to those derived from landmark estimation methods, leading to local optima during the image based optimization. While landmark-based registration \cite{Grupp2020-xz,Bier2018-ht} typically provides more accurate initial estimates, it faces several challenges. These include the requirement for 3D landmark annotation during inference and failures in views with insufficient landmark visibility. One approach to address these issues involves backprojecting 2D landmark estimates from multi-view DRRs and fine-tuning the landmark estimation model with synthetic images \cite{Esteban2019-zg}. Recent study have shown the effectiveness of training deep models with synthetic images for vision tasks related to X-ray images \cite{Gao2023-ep}. Additionally, scene coordinates can be directly estimated from X-ray images \cite{Shrestha2023-nk}, eliminating the need for 3D landmark annotation. Although effective, these patient-specific methods are less suitable for urgent scenarios, such as emergency surgeries, due to their preparatory requirements.

\begin{figure}[tb]
  \centering
  \includegraphics[width=1.0\textwidth]{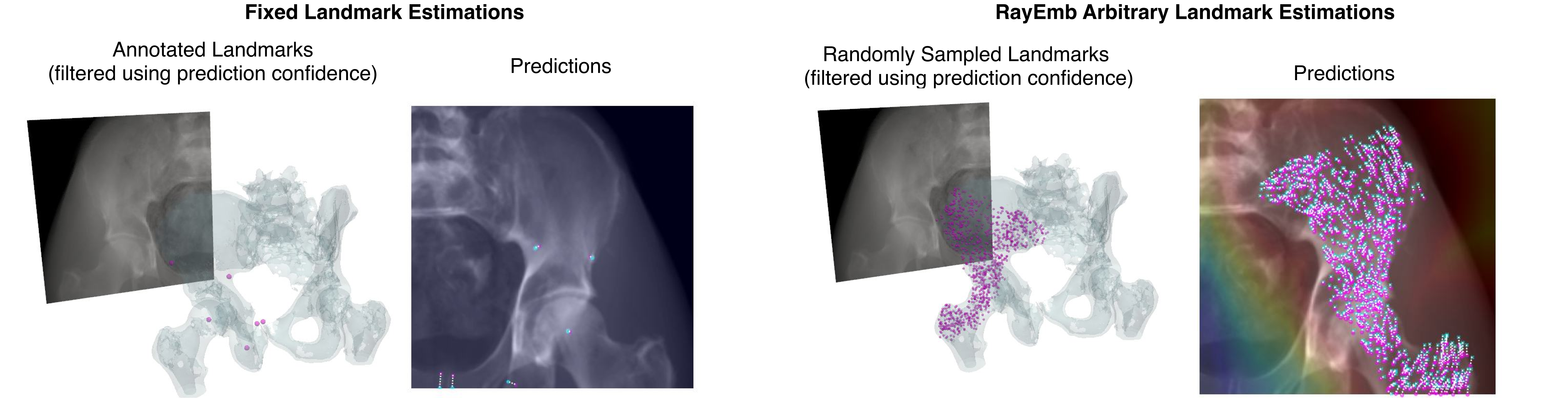}
  \caption{
Comparison of landmark detection results between conventional fixed landmark estimation and our arbitrary landmark estimation method. The 3D landmarks are shown in magenta on the left, while the estimated 2D landmarks are displayed in cyan and the ground truth in magenta on the right. Our method can generate a large number of corresponding pairs of 3D landmarks and 2D projections, whereas the fixed landmark estimation approach is limited to the available 2D projections and does not provide 3D landmark information.
  }
  \label{fig:concept}
\end{figure}

\section{Methods}
At a high level, our proposed method involves arbitrary landmark estimation followed by 2D-3D registration. Fig. \ref{fig:overview} illustrates an overview of our landmark estimation process. Once the correspondences are established, the perspective-n-point algorithm \cite{Lu2018-tx} with marginalizing sample and consensus (MAGSAC) \cite{Barath2019-ns} is used to obtain the initial pose estimate. This estimate serves as initialization for DiffDRR \cite{Gopalakrishnan2023-re}, a gradient-based optimization refinement module. The core idea is to provide pose estimator with large amount of corresponding pairs of 3D landmarks and its 2D projections. \cref{fig:concept} compares this approach to the conventional fixed landmark estimation method. 

\subsection{Problem Formulation}
The X-ray image formation is typically described using Beer-Lambert's Law\cite{Swinehart1962-nb}. 
\begin{equation}
\label{eq:1}
I(\mathbf{d}; \mathbf{T}) = I_0 \exp \left( - \int_{0}^{x} \mu(t\mathbf{d}; \mathbf{T}) \, dt \right)
\end{equation}
where \( I(\mathbf{d} \) is the intensity of the X-ray after passing through the material. \( I_0 \) is the initial intensity of the X-ray before entering the material, \( \mu \) is the linear attenuation coefficient of the volume, \( \mathbf{d} \) is the direction vector towards the detector pixel in camera space, and \(\mathbf{T}\) represents the pose of the volume. Given a target X-ray image \(I_{target}\), the goal is to find the optimal pose \(\mathbf{T}_{optim}\) that minimizes the discrepancy between the rendered image and the target image.
\begin{equation}
\label{eq:2}
\mathbf{T}_{optim} = \argmin_{\mathbf{T}} d(I_{target}, I(\mathbf{T}))
\end{equation}
where \(d\) measures the distance between the two images. In practice, effectively solving \cref{eq:2} requires an accurate initial pose estimate \(T\) to ensure convergence to the correct solution. We propose to obtain a suitable initialization for this pose by establishing 2D-3D correspondences. However, as evident from \cref{eq:1}, establishing direct correspondence between a pixel and a 3D point is challenging, since the pixel intensity represents the integral of attenuation coefficients along the ray path. In other words, a many-to-one correspondence must be established between the points along the ray and the pixel. 

\subsection{Arbitrary Landmark Estimation}
Estimating correspondences for arbitrary landmark points between 2D and 3D representations poses significant challenges due to the mismatch in spatial dimensions and the transmissive properties of X-rays, which can associate multiple 3D points with a single 2D projection. To address this, we employ a pixel-wise feature extractor and pre-render DRR templates to represent a 3D point as a subspace. The pixel-wise features are referred to as ray embeddings, as we associate the feature vector to its back-projected ray. Since a 3D point can be represented by a collection of intersecting rays, a set of ray embeddings can be associated with a 3D point if their underlying rays intersect. The 2D projection of a 3D point can then be identified by evaluating the closeness of the ray embeddings in the query image to the subspace representing the 3D point. The following sections will provide a detailed explanation of the main components of the proposed method.

\subsubsection{Ray Embeddings.}
The input query X-ray image and the pre-rendered templates are both fed into an encoder for obtaining pixel wise ray embedding vectors. Formally, let \(I_{q}(\mathbf{x})\) be the input query image and \(I^{'}_{t}(\mathbf{x})\) be the template images, where \(t \in \{1, 2, 3, ..., N\} \), with \(N\) being the number of templates. The ray embedding for each image point \(\mathbf{x}\) in query input image is defined as \( \mathbf{e}(\mathbf{x}) = E(I_{q}(\mathbf{x}); \mathbf{w})\), where \( E: \mathbb{R} \to \mathbb{R}^d\). Similarly, the ray embedding in the template image is given by \( \mathbf{e'_{t}}(\mathbf{x}) = E(I'_{t}(\mathbf{x}); \mathbf{w})\). It should be noted that both the query image and template images are processed using the same encoder to ensure that the resulting embedding vectors reside in the same contextual space.

\subsubsection{Subspace Representation of 3D Landmarks.}
Let \(\mathbf{X}\) represent a sampled 3D landmark inside the CT volume (as shown in the left section of Fig. \ref{fig:overview}). Given the known camera transformation matrices \(\mathbf{T'_{t}}\) for each template image, we can compute the projection of \(\mathbf{X}\) onto the template image planes as \( \mathbf{x'_{t}} =  \pi_\mathbf{K} (\mathbf{X}; \mathbf{T'_{t}})\), where \(\pi_\mathbf{K} (\cdot)\) is the camera projection operator. For clarity, the ray embeddings at these projected points are denoted as follows:
\begin{equation}
\label{eq:3}
    \mathbf{e'_{t}} = E(I'_{t}(\mathbf{x'_{t}}); \mathbf{w}),\quad t \in \{1, 2, 3, ..., N\}
\end{equation}
The embedding vectors are then stacked to form a transformation matrix \(\mathbf{F}\), whose column space is spanned by the embedding vectors of rays that intersect at \(\mathbf{X}\) in 3D space.
\begin{equation}
\label{eq:4}
    \mathbf{F} = (\mathbf{e'_{1}}, \mathbf{e'_{2}}, ..., \mathbf{e'_{N}})
\end{equation}

The orthogonal projection \( \mathbf{P}\) onto the subspace spanned by the embedding vectors can be calculated by  right multiplying \(\mathbf{F}\) with its Moore-Penrose inverse \(\mathbf{F^{+}} = \mathbf{V} \mathbf{\Sigma^{+}} \mathbf{U^{T}}\), where \(\mathbf{U}, \mathbf{\Sigma}\), and \(\mathbf{V}\) are the matrices containing the left singular vectors, the singular values, and the right singular vectors of \(\mathbf{F}\), respectively. 

\begin{equation}
\label{eq:5}
    \mathbf{P} = \mathbf{F} \mathbf{F^{+}}  = \mathbf{U} \mathbf{\Sigma} \mathbf{\Sigma^{+}} \mathbf{U^{T}}
\end{equation}

The projection matrix \(\mathbf{P}\) now represents an \(N\)-dimensional subspace within a  \(D\)-dimensional embedding space. We associate this projection matrix with the 3D point \(\mathbf{X}\).

\subsubsection{Corresponding 2D Landmark Estimation.}
During inference, ray embeddings of the query input image are available for all spatial grid points on the detector plane. For a 3D point \(\mathbf{X}\), we calculate the cosine similarity between the projection of each ray embedding onto the subspace and the embedding itself:
\begin{equation}
\label{eq:6}
    \text{sim}(\mathbf{P}, \mathbf{x}) = \frac{\mathbf{e^{T}(\mathbf{x})} \mathbf{P} \mathbf{e(\mathbf{x})}}{\left|\mathbf{e^{T}(\mathbf{x})}\right| \left|\mathbf{P} \mathbf{e(\mathbf{x})}\right|}
\end{equation}
where \(\mathbf{x}\) represents 2D points on the detector plane. The corresponding 2D point \(\mathbf{\hat{x}}\) for the given 3D point \(\mathbf{X}\), associated with subspace \(\mathbf{P}\), is identified by finding the location with maximum similarity value across the detector plane.
\begin{equation}
\label{eq:7}
\mathbf{\hat{x}} = \argmax_{\mathbf{x}} \text{sim}(\mathbf{P}, \mathbf{x})
\end{equation}

\subsubsection{Contrastive Learning.}
The goal is to train the encoder to learn embeddings such that rays intersecting at the same 3D point are mapped to a distinct subspace. During training, we have ray embedding vectors from the templates that intersect in 3D space, along with a positive sample \(\mathbf{x^{+}}\) from the query image that also intersects the same 3D point. All other embedding vectors are treated as negative samples \(\mathbf{x^{-}}\). The InfoNCE loss \cite{van-den-Oord2018-kz} is employed to achieve this:
\begin{equation}
\label{eq:8}
\mathcal{L} (\mathbf{e}, \mathbf{e'_t})= -\log \frac{\exp(\text{sim}(\mathbf{P}, \mathbf{x^{+}}) / \tau)}{\sum_{\mathbf{x^-} \in \mathcal{N}} \exp(\text{sim}(\mathbf{P}, \mathbf{x^-}) / \tau)}
\end{equation}
Here, \(\tau\) is the temperature constant, and \(\mathcal{N}\) includes the positive sample as well.

\subsection{2D-3D Registration}
Using the 2D-3D correspondences and similarity scores obtained from Equation \ref{eq:6}, we adopt MAGSAC \cite{Barath2019-ns}, a variant of RANSAC that does not require an explicitly set inlier threshold, to compute the initial pose estimate. Although MAGSAC is capable of rejecting outlier correspondences that arise from points outside the field of view, we pre-filter the correspondences by selecting the top-\(k\) points with the highest similarity values. This filtering step ensures that the resulting correspondences are mostly within the field of view, as shown in \cref{fig:concept}. To further improve robustness, we apply test-time augmentation during subspace generation by randomly choosing the templates \(n\)-times and selecting the 2D projections with the maximum similarity response. Following this, a gradient-based optimization of the pose is performed using multi-scale normalized cross-correlation with differentiable rendering, in line with the approach of DiffPose \cite{Gopalakrishnan2024-yq}.

\begin{figure}[t]
\centering
\includegraphics[width=1.0\linewidth]{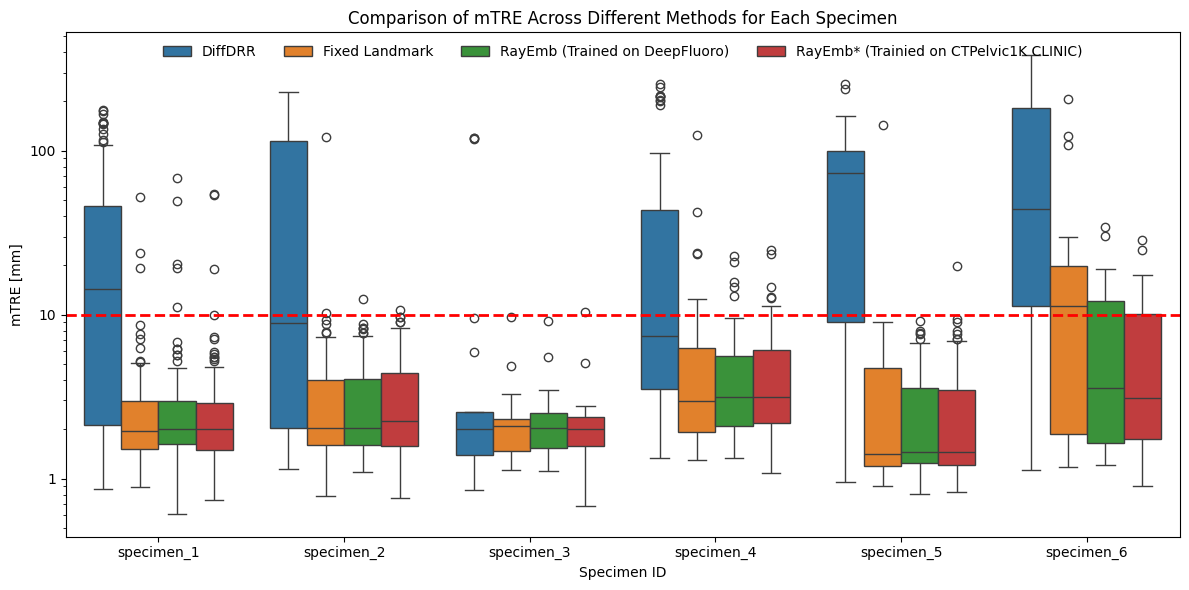}
\caption{
Box plots comparing the mean Target Registration Errors (mTRE) for six specimens using three different registration methods: DiffDRR, Fixed Landmark, and our proposed method, RayEmb (trained on DeepFluoro) and RayEmb* (trained on CTPelvic1K CLINIC). A red dashed line at 10 mm indicates the critical threshold for acceptable performance limit.}
\label{fig:main_mTRE}
\end{figure}

\section{Experiments}

\subsection{Datasets}
Following prior research, we utilized the DeepFluoro dataset \cite{Grupp2020-xz} to evaluate our proposed method. Although we generated synthetic images using the CT volumes for training, we evaluated the model on real X-rays from this dataset. Additionally, we trained the model on the CTPelvic1K CLINIC\cite{Liu2021-kf} dataset to assess its generalization capability when tested on real X-ray image from DeepFluoro dataset. 

\subsubsection{DeepFluoro.} This dataset comprises CT scans from six specimens, totaling 366 real X-ray images registered semi-automatically. For each specimen, 14 anatomical landmarks are labeled in the CT volume and their corresponding projections on the real X-ray images. The X-ray images were collected using a Siemens CIOS Fusion mobile C-arm device with \(30 \times 30\) \(cm^{2}\) detector size, \(1536 \times 1536\) pixels with 0.194 mm pixel spacing. The 3D anatomical landmarks were used to train the baseline model; however, these landmarks were not required, and therefore not used, during training or inference with our proposed method. We used DiffDRR \cite{Gopalakrishnan2024-yq} to generate 10,000 synthetic images for each specimen by sampling random poses, simulating the C-arm CIOS Fusion device. Synthetic images from specimens 1 to 5 were used for training and validation, while specimen 6 was reserved exclusively for testing.

\subsubsection{CTPelvic1K CLINIC.}
To evaluate the generalizability of the proposed method across different patient CT scans, we also trained our model using the CTPelvic1K CLINIC dataset, which is a subset comprising 103 CT scans. This dataset is part of a larger collection originally containing approximately 1184 CT volumes with varying resolutions and a diverse range of subjects, each annotated with pelvic segmentation labels. For this study, we specifically selected the CLINIC subset, which provides a balanced sample size, avoiding extremes of very large or very small numbers of CT volumes, thereby offering a representative sample for effective training. Similar to the DeepFluoro dataset, we used DiffDRR to generate 1000 synthetic images per CT scan, doubling the total number of images than that of the DeepFluoro dataset. The model trained on the CTPelvic1K CLINIC dataset was then evaluated on real X-ray images from the DeepFluoro dataset.

\subsection{Baselines and Evaluation Metrics}
We compare the RayEmb method against a conventional approach that identifies fixed spatial landmarks, referred to here as Fixed Landmark. The Fixed Landmark method estimates heatmaps of 14 anatomical landmarks using a U-Net-based network, following the architecture described in \cite{Grupp2020-xz} but without the segmentation module, instead regressing heatmaps at the decoder output. Since both methods incorporate pose refinement using DiffDRR, we also compare them against DiffDRR initialized from a standardized anterior-posterior (AP) view to ensure consistency across evaluations. We report two key metrics for registration: mean target registration error (mTRE) and mean projected distance (mPD), defined as follows:
\begin{equation}
\label{eq:9}
  \text{mTRE}(\mathbf{T}, \mathbf{\hat{T}}) = \frac{1}{N} \sum_{i=1}^N \| \mathbf{T}\mathbf{X}_i  - \mathbf{\hat{T}} \mathbf{X}_i \|
\end{equation}
\begin{equation}
\label{eq:10}
  \text{mPD}(\mathbf{T}, \mathbf{\hat{T}}) = \frac{1}{N} \sum_{i=1}^N \| \pi_\mathbf{K}(\mathbf{X_i}; \mathbf{T}) - \pi_\mathbf{K}(\mathbf{X_i}; \mathbf{\hat{T}})\|
\end{equation}
where \( \mathbf{X}_i \) is the 3D landmark, \(\mathbf{T}\mathbf{X_i}\) represents landmark in camera space, and \( \pi_\mathbf{K}(\mathbf{X_i}; \mathbf{T})\) denotes its projection onto the detector plane. Both mTRE and mPD are measured in millimeters. Additionally, we report the gross failure rate (GFR), defined as the proportion of images that could not be registered within a specified tolerance threshold (10mm or 5mm), for each specimen.

\begin{table}[tb]
  \caption{The mTRE for each 25th, 50th, and 95th percentile of the DeepFluoro dataset, along with GFR at 10mm and 5mm in percentage, are reported. The proposed method trained on DeepFluoro as well as CTPelvic1K achieves the lowest errors in all metrics for specimen 6, which includes challenging views.}
  \label{tab:mTRE}
  \centering
\scriptsize
   \begin{tabular}{@{}c>{\raggedright\arraybackslash}m{3cm}llllll@{}}
    \toprule
    \textbf{Specimen} & \textbf{Method} & \textbf{\boldmath\(25^{th}\)} & \textbf{\boldmath\(50^{th}\)} & \textbf{\boldmath\(95^{th}\)} & \textbf{GFR@10\(\downarrow\)} & \textbf{GFR@5\(\downarrow\)} \\
    \midrule
    \multirow{4}{*}{1} & DiffDRR & 2.125 & 14.259 & 147.349 & 54.054 & 63.964 \\
                       & Fixed Landmark & 1.520 & \textbf{1.960} & 6.701 & \textbf{2.703} & \textbf{9.009} \\
                       & RayEmb & 1.625 & 2.010 & \textbf{6.520} & 4.505 & 9.910 \\
                       & RayEmb* & \textbf{1.502} & 2.011 & 6.535 & \textbf{2.703} & 10.811 \\
    \midrule
    \multirow{4}{*}{2} & DiffDRR & 2.049 & 8.879 & 203.835 & 48.077 & 59.615 \\
                       & Fixed Landmark & \textbf{1.602} & \textbf{2.027} & \textbf{7.660} & 1.923 & 19.231 \\
                       & RayEmb & 1.604 & 2.049 & 7.772 & \textbf{0.962} & \textbf{17.308} \\
                       & RayEmb* & 1.578 & 2.264 & 8.281 & \textbf{0.962} & 20.192 \\
    \midrule
    \multirow{4}{*}{3} & DiffDRR & \textbf{1.402} & 2.007 & 119.635 & 12.500 & 20.833 \\
                       & Fixed Landmark & 1.474 & 2.100 & \textbf{4.628} & \textbf{0.000} & \textbf{4.167} \\
                       & RayEmb & 1.549 & 2.053 & 5.239 & \textbf{0.000} & 8.333 \\
                       & RayEmb* & 1.583 & \textbf{2.019} & 4.737 & 4.167 & 8.333 \\
    \midrule
    \multirow{4}{*}{4} & DiffDRR & 3.529 & 7.402 & 216.490 & 41.667 & 66.667 \\
                       & Fixed Landmark & \textbf{1.937} & \textbf{2.965} & 23.770 & 14.583 & 35.417 \\
                       & RayEmb & 2.094 & 3.157 & 15.375 & \textbf{10.417} & \textbf{29.167} \\
                       & RayEmb* & 2.186 & 3.167 & \textbf{14.103} & 12.500 & 33.333 \\
    \midrule
    \multirow{4}{*}{5} & DiffDRR & 9.062 & 73.597 & 140.134 & 74.545 & 76.364 \\
                       & Fixed Landmark & \textbf{1.190} & \textbf{1.414} & 8.172 & 1.818 & 25.455 \\
                       & RayEmb & 1.248 & 1.447 & \textbf{7.633} & \textbf{0.000} & \textbf{18.182} \\
                       & RayEmb* & 1.211 & 1.452 & 8.313 & 1.818 & 21.818 \\
    \midrule
    \multirow{4}{*}{6} & DiffDRR & 11.263 & 43.853 & 375.553 & 75.000 & 79.167 \\
                       & Fixed Landmark & 1.875 & 11.250 & 121.944 & 50.000 & 62.500 \\
                       & RayEmb & \textbf{1.647} & 3.580 & 28.496 & 33.333 & \textbf{45.833} \\
                       & RayEmb* & 1.754 & \textbf{3.090} & \textbf{23.832} & \textbf{25.000} & \textbf{45.833} \\
    \bottomrule
  \end{tabular}
\end{table}

\begin{table}[tb]
  \caption{The mPD for each 25th, 50th, and 95th percentile of the DeepFluoro dataset, along with GFR at 10mm and 5mm threshold in percentage, are reported. Best values highlighted for each metric across specimens.}
  \label{tab:mPD}
  \centering
\scriptsize
  \begin{tabular}{@{}c>{\centering\arraybackslash}m{3cm}llllll@{}}
    \toprule
    \textbf{Specimen} & \textbf{Method} & \textbf{\boldmath\(25^{th}\)} & \textbf{\boldmath\(50^{th}\)} & \textbf{\boldmath\(95^{th}\)} & \textbf{GFR@10\(\downarrow\)} & \textbf{GFR@5\(\downarrow\)} \\
    \midrule
    \multirow{4}{*}{1} & DiffDRR & 1.681 & 5.476 & 172.835 & 45.045 & 52.252 \\
                       & Fixed Landmark & 1.417 & 1.696 & \textbf{2.598} & \textbf{1.802} & \textbf{1.802} \\
                       & RayEmb & 1.420 & 1.722 & 2.676 & \textbf{1.802} & 3.604 \\
                       & RayEmb* & \textbf{1.340} & \textbf{1.676} & \textbf{2.636} & \textbf{1.802} & 2.703 \\
    \midrule
    \multirow{4}{*}{2} & DiffDRR & 1.868 & 2.877 & 240.662 & 45.192 & 47.115 \\
                       & Fixed Landmark & 1.589 & 1.808 & \textbf{2.692} & 0.962 & 0.962 \\
                       & RayEmb & 1.610 & 1.817 & 2.741 & \textbf{0.000} & \textbf{0.000} \\
                       & RayEmb* & \textbf{1.515} & \textbf{1.793} & 2.852 & \textbf{0.000} & \textbf{0.000} \\
    \midrule
    \multirow{4}{*}{3} & DiffDRR & 1.311 & 1.722 & 166.758 & 12.500 & 12.500 \\
                       & Fixed Landmark & 1.653 & 1.739 & 2.962 & \textbf{0.000} & \textbf{0.000} \\
                       & RayEmb & 1.523 & \textbf{1.824} & \textbf{2.950} & \textbf{0.000} & \textbf{0.000} \\
                       & RayEmb* & \textbf{1.393} & 1.846 & 3.313 & \textbf{0.000} & \textbf{0.000} \\
    \midrule
    \multirow{4}{*}{4} & DiffDRR & 2.984 & 5.100 & 290.629 & 35.417 & 50.000 \\
                       & Fixed Landmark & 2.146 & \textbf{2.519} & 7.091 & 2.083 & 10.417 \\
                       & RayEmb & \textbf{2.046} & 2.572 & \textbf{4.755} & \textbf{0.000} & \textbf{4.167} \\
                       & RayEmb* & 2.314 & 2.783 & 6.662 & \textbf{0.000} & 14.583 \\
    \midrule
    \multirow{4}{*}{5} & DiffDRR & 2.405 & 56.259 & 171.329 & 70.909 & 70.909 \\
                       & Fixed Landmark & 1.295 & \textbf{1.536} & 2.654 & 1.818 & 1.818 \\
                       & RayEmb & \textbf{1.288} & 1.561 & \textbf{2.152} & \textbf{0.000} & \textbf{0.000} \\
                       & RayEmb* & 1.355 & 1.630 & 2.705 & \textbf{0.000} & \textbf{0.000} \\
    \midrule
    \multirow{4}{*}{6} & DiffDRR & 3.232 & 22.468 & 257.309 & 62.500 & 66.667 \\
                       & Fixed Landmark & 2.251 & 3.433 & 152.635 & 20.833 & 20.833 \\
                       & RayEmb & 2.006 & \textbf{2.567} & 6.096 & \textbf{0.000} & 20.833 \\
                       & RayEmb* & \textbf{1.838} & 2.605 & \textbf{5.093} & \textbf{0.000} & \textbf{8.333} \\
    \bottomrule
  \end{tabular}
\end{table}

\subsection{Implementation Details}
\subsubsection{Training.}
We use a DINOv2 \cite{Oquab2023-ve} pretrained Vision Transformer encoder with a fully connected layer to convert 768-dimensional descriptors into 32-dimensional embedding vectors. Each input image is resized to \( 224 \times 224 \) pixels. A total of 324 template DRRs are generated from equally spaced pose samples, covering 45 degrees in the LAO/RAO directions and 22.5 degrees in the CRA/CAU directions, with 18 steps taken in each direction. During training, 4 templates are randomly selected to generate subspaces for the given sampling points. PyTorch is used to compute the backpropagatable pseudo-inverse, and gradient clipping with a maximum value of 10.0 is applied. The model is trained using the Adam optimizer with a learning rate of 1e-4 for 50 epochs on an RTX 3090 GPU Ti, with a batch size of 8, 40 3D sampling points, and a temperature parameter of 1e-4. The Fixed Landmark method is also trained using the same batch size and optimizer settings.

\subsubsection{Inference.}
We use the PnP and RANSAC implementation from OpenCV for initial registration. Specifically, we sample 3000 points inside the bone segmentation mask of the CT volume, run our pipeline, and select top-\(k\) points based on similarity scores for initial filtering, with \(k = 600 \) in our experiments. For these top-\(k\) points, we run the pipeline n-times (\(n=10\)), using randomly sampled templates in each iteration, and choose the 2D projection with the highest similarity values. It should be noted that these hyperparameters were manually chosen based on testing with DeepFluoro synthetic validation cases. For the refinement procedure, we follow the same protocol as in DiffPose \cite{Gopalakrishnan2024-yq}, running the optimization for 100 iterations for both the proposed method and the Fixed Landmark method. To compare with DiffDRR-only method, we initialize the pose with AP view for each specimen and run optimization for 150 iterations.

\begin{figure}[t]
\centering
\includegraphics[width=1.0\linewidth]{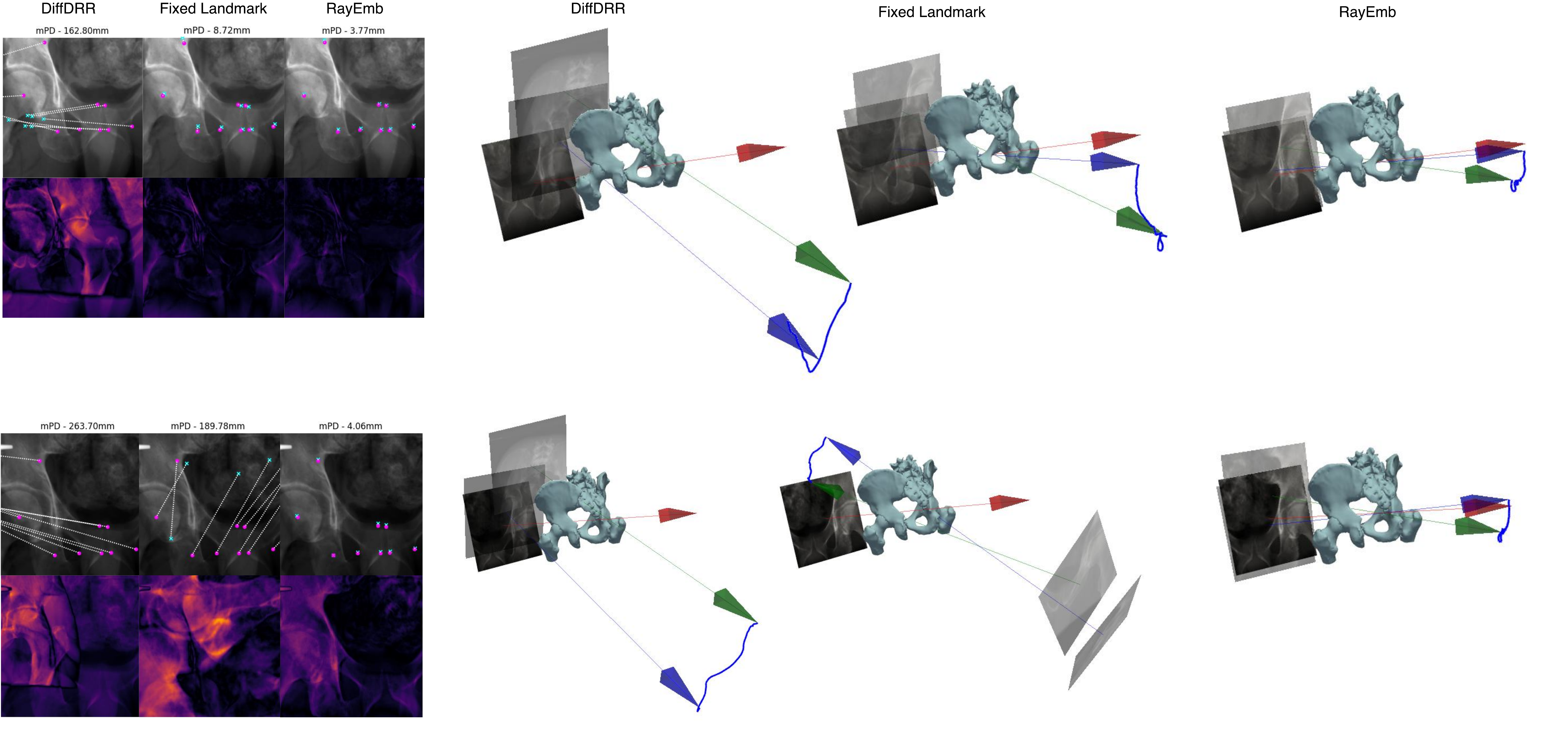}
\caption{Two example cases (Top and Bottom) where RayEmb provides better initial pose estimates compared to other methods. The ground truth landmarks are shown in magenta, and the estimated landmarks are in cyan (left). A 3D visualization of the initial pose estimate is displayed in green, the optimized pose in blue, and the ground truth pose in red (right).}
\label{fig:comapare}
\end{figure}

\subsection{Registration Results}
\subsubsection{Quantitative.}
We evaluated the mean Target Registration Errors (mTRE) across three registration methods $-$ DiffDRR, Fixed Landmark, RayEmb $-$ for six different specimens. RayEmb refers to the proposed method trained on DeepFluoro dataset, while RayEmb* indicates the method trained on the CTPelvic1K CLINIC dataset. The results are presented as box plots in Fig. \ref{fig:main_mTRE}. DiffDRR exhibited the highest median errors and variability across all specimens. In particular, specimens 2, 4, and 6 showed significantly higher mTREs with this method, far exceeding the 10 mm threshold, suggesting limitation in initializing an image-based similarity optimization scheme with a constant AP pose for achieving precise registration. Table. \ref{tab:mTRE} and Table. \ref{tab:mPD} show that both RayEmb and Fixed Landmark maintain median mTREs below the 10 mm and mPD below 5mm threshold across all specimens. The mPD values are generally lower than mTRE because mPD is less sensitive to error in the depth direction.  Notably, for specimen 6, which presents a challenging test case due to patient and pose variability, the proposed method achieves lower mTRE, mPD, and GFR, suggesting better generalizability to unseen patient CTs.

\subsubsection{Qualitative.}
Fig. \ref{fig:comapare} shows two example test cases from specimen 6 where initial pose estimation is crucial for successful pose refinement. In the left section, the ground truth landmarks are shown in magenta, while the estimated landmarks from each method are indicated in cyan. The top case demonstrates the proposed method providing a pose estimate closer to the ground truth than the Fixed Landmark method, while the bottom case illustrates the proposed method's ability to generate pose estimates even in situations where both DiffDRR and Fixed Landmark fail. RayEmb achieves significantly closer alignment to the ground truth, with mPD values of 3.77 mm and 4.06 mm, which are substantially lower than those of DiffDRR and Fixed Landmark. In the right section, the initial pose estimates are shown in green, while the ground truth poses are displayed in red. The trajectories between the estimated and true poses illustrate the deviation of each method from the ideal alignment. RayEmb shows smaller deviations compared to the other methods, highlighting its superior precision in initial pose estimation. Additional examples are provided in the supplemental material.

\begin{figure}[t]
\centering
\includegraphics[width=1.0\linewidth]{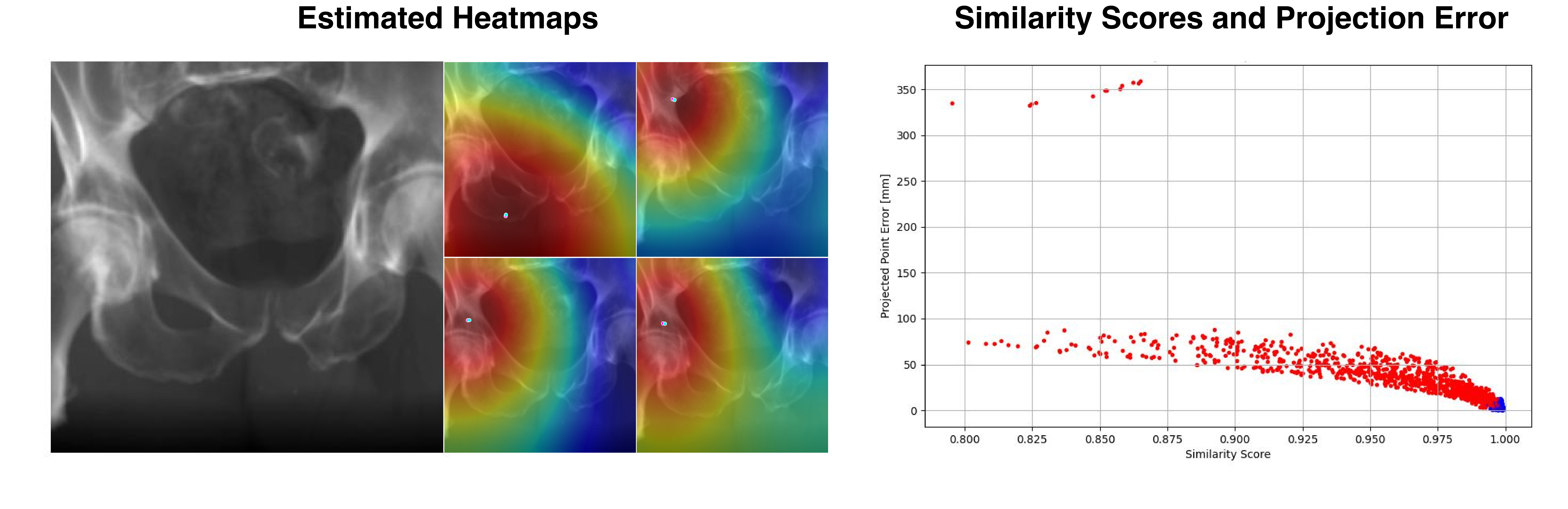}
\caption{Examples of the estimated heatmaps (left) and the similarity score versus projection error plot (right). In the heatmaps, the cyan point indicates the predicted point, while the magenta point represents the corresponding ground truth. In the plot, red points correspond to sampling points outside the field of view, and blue points correspond to visible sampling points.}
\label{fig:analyze}
\end{figure}

\subsection{Analyzing the Heatmap and Similarity Scores}
\cref{fig:analyze} presents estimated heatmaps alongside a plot of similarity scores versus projection errors, derived from the DeepFluoro synthetic dataset. The heatmaps demonstrate RayEmb's ability to accurately estimate landmark positions. In the plot, red points represent sampling points outside the field of view, while blue points indicate visible sampling points within the projected image field. This distinction is crucial as it provides insights into the spatial distribution and accuracy of the landmark estimations relative to the actual imaging area. The plot shows a strong correlation between high similarity scores and low projection errors, validating the effectiveness of selecting top-\(k\) correspondences based on high similarity metrics. By prioritizing correspondences with the highest similarity scores, RayEmb effectively minimizes errors in subsequent steps, thereby enhancing both the accuracy and reliability of the 2D-3D registration process.

\section{Limitations}
\begin{figure}[t]
\centering
\includegraphics[width=0.5\linewidth]{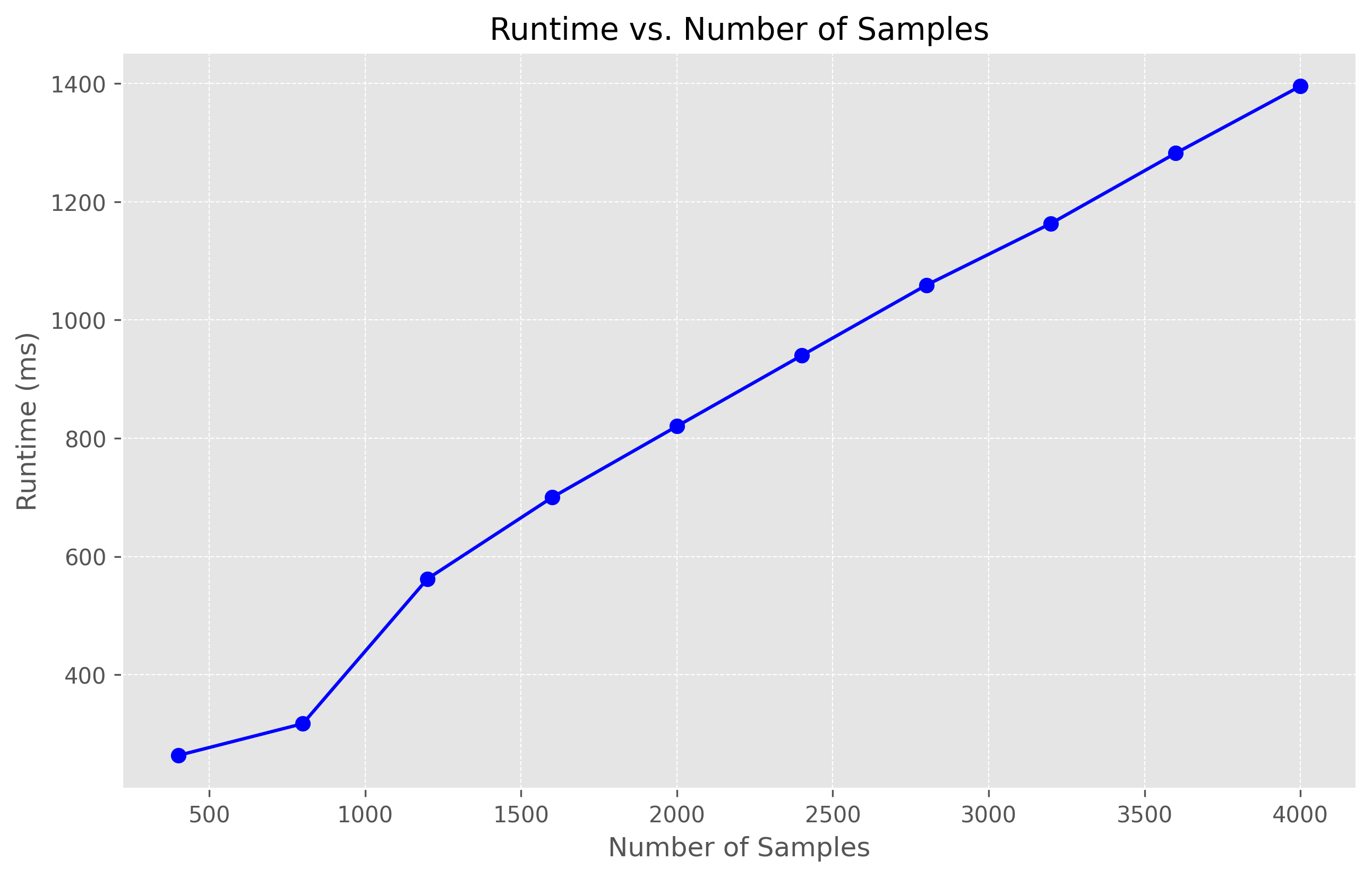}
\caption{Inference runtime of the proposed method, excluding the DiffDRR optimization.}
\label{fig:runtime}
\end{figure}

While the proposed method achieves precise registration with enhanced generalizability, the runtime of the landmark estimation pipeline depends on the number of sampled points; a higher number of samples leads to increased computation time. As shown in Fig. \ref{fig:runtime}, current inference setup with 3000 sampling points require approximately 1000ms for initial pass and another 2000ms (\(200ms \times 10\)) for test time augmentation, excluding the DiffDRR optimization, on an RTX 3090 GPU Ti. Additionally, all three methods employ DiffDRR with a multi-scale normalized cross-correlation loss, which requires few seconds to complete 100 iterations. The efficiency related to sampling points could be improved by implementing hierarchical sampling and filtering, which would use similarity scores to exclude points outside the specific region of interest for a given view. Furthermore, the refinement process could be significantly accelerated by adopting a sparse implementation of normalized cross-correlation (sparse mNCC), as used in DiffPose.

\section{Conclusion}
We demonstrated the effectiveness of arbitrary landmark detection using the ray embedding subspace method to achieve high-precision registration across various specimens, with notable success in the challenging test case of specimen 6. Additionally, we validated the generalizability of the proposed method by training on the CTPelvic1K dataset and evaluating on the DeepFluoro dataset, resulting in a patient-wide model that performs well without any patient-specific fine-tuning steps required by previous approaches. This approach addresses critical needs in emergency treatments and diagnostics, where both accuracy and broad generalizability across patients are essential. Future work will focus on further refining the method and exploring its applicability to other anatomical regions and clinical scenarios.

\subsubsection{Acknowledgments.}
This work was partially supported by a grant from JSPS KAKENHI Grant Number JP23K08618. This work used computational resources of Pegasus provided by Multidisciplinary Cooperative Research Program in Center for Computational Sciences, University of Tsukuba.

\bibliographystyle{splncs04}
\bibliography{main}
\end{document}